%% file: paper.tex
\ifdefined\anonymousversion
  \PassOptionsToClass{anonymous}{acmart}
\fi
\documentclass[sigconf,nonacm]{acmart}
\usepackage{amsmath}
\usepackage{booktabs}
\usepackage{longtable}
\usepackage{pgfplots}
\pgfplotsset{compat=1.18}
\setcopyright{none}
\input{generated/numbers}
\input{generated/curve_data}
\input{generated/extended_numbers}
\input{generated/local_model_numbers}
\newcommand{\curve}[2]{%
  \edef\plotcommand{\noexpand\addplot+[mark size=1.8pt]
    coordinates {\csname #1#2\endcsname};}\plotcommand}

\title{What Structurally Breaks an Agent?\\
Auditing Difficulty in Stateful Tool-Use Workflows}
\ifdefined\anonymousversion
\author{Anonymous Author(s)}
\else
\author{Shereen Anand}
\email{pes1202201328@pesu.pes.edu}
\affiliation{\institution{PES University}\country{India}}
\author{Ju Wang}
\email{jun.wang@cs.ucl.ac.uk}
\affiliation{\institution{University College London}
  \country{United Kingdom}}
\author{Yali Du}
\email{yali.du@kcl.ac.uk}
\affiliation{\institution{King's College London}
  \country{United Kingdom}}
\fi

\begin{abstract}
An aggregate task score does not distinguish failures of scheduling, state
retention, and candidate selection. We develop a small executable instrument
that varies dependency depth, candidate count, active state width, and access
to current-state snapshots while holding the number of required operations
fixed. Two dependency-graph families expose an important limitation of
structural control: increasing depth at fixed work changes both scheduling
freedom and the number of reachable completion states. We quantify these
couplings instead of treating the controls as universally independent.
The primary study contains \TaskCount{} generated tasks and \EpisodeCount{}
reference-controller rollouts. A separately frozen follow-on design adds
\ExtendedEpisodeCount{} rollouts covering operation count, retention
strategy, snapshot cadence, hidden state resampling, recovery, query locality,
update removal, candidate indexing, redundant computation, and relabeling.
The ablations expose representation dependence: residue caches and precomputed
plans solve tasks that defeat limited raw-value caches, while indexed
selection removes the serial-scan branching deficit. Hidden resampling
instead imposes an exact success ceiling that additional state tracking
cannot overcome. Retry-enabled elimination changes the apparent failure
threshold at an explicit tool-call cost. The contribution is an executable
diagnostic instrument, analytic calibration, and an audit of its assumptions;
a small, single-model local CPU study adds an exploratory model-in-the-loop
check, but its 256 tasks yielded no completed workflows for the 0.5B model.
This floor result does not show that the task dimensions have no effect and
supports no cross-provider or reasoning-budget claim. A separate 16-episode,
two-seed fractional factorial with one Copilot-routed model completed half
the tasks; its small sample and aliased interactions make the contrasts
descriptive rather than confirmatory.
\end{abstract}

\begin{document}
\maketitle
\fancyhf{}
\fancyfoot[C]{\thepage}
\renewcommand{\headrulewidth}{0pt}
\raggedbottom

\section{Introduction}\footnote{\href{https://github.com/Shereen20112003/what-structurally-breaks-an-agent}{GitHub}.}
What makes a task difficult for an agent? A long sequence may require little
planning if each next action is obvious. A short task may fail because a
necessary value is no longer visible. Many available actions need not be
problematic when their preconditions are easy to evaluate. These cases have
different remedies, yet a single completion score treats them alike.

Existing work already establishes that task complexity matters.
DyVal generates reasoning problems with controllable graph structure
\cite{dyval}; controlled puzzles examine performance as compositional complexity
increases \cite{illusion}; and task depth and capability breadth have been
studied in multi-agent evaluation \cite{tang}. More directly, TRAJECT-Bench
varies tool-trajectory breadth and depth and reports trajectory-level
selection, argument, and ordering metrics \cite{traject}. The Reward Hacking
Benchmark studies independent and chained multi-step tool tasks across
frontier models, including harder task variants \cite{rhb}. SynthAPI varies
tool, task, and database diversity in synthetic executable environments,
although its question is training-data scaling rather than capability
diagnosis \cite{synthapi}. Synthetic POMDPs have also been designed with explicit, theoretically
characterized memory-demand structures while preserving transition
invariants \cite{mds}. Our remaining question is narrower: can a small
tool-execution workflow isolate state-retention and feedback conditions
while preserving required work and auditing unavoidable structural
couplings?

The scope of novelty is necessarily narrow. Controllable graph difficulty,
reasoning-budget comparisons, tool-trajectory diagnostics, multi-step tool
tasks, and memory benchmarks all have substantial prior work
\cite{dyval,illusion,tokenbudget,apibank,toolbench,appworld,mcpagentbench,
traject,rhb,synthapi,memcalib}. Our candidate distinction is the combination
of a fixed-work, state-mutating workflow with explicit contrasts between
state snapshots and history-reconstructible feedback, plus audits of how
controller representation changes the resulting profile. This is a
testable distinction, not a novelty finding: neither the reference
controllers nor the two small, single-model evaluations validate a general
capability-profile contribution, and no direct benchmark comparison has been
performed.

We study four operational controls: the longest prerequisite chain,
the number of candidates within an operation, the number of active state
registers, and whether current register values are repeated in feedback.
The total number of required operations is always eight. This is a
\emph{fixed-work depth intervention}: depth changes precedence, not episode
length. Candidate count is likewise local to an operation, not the number
of globally available actions. These distinctions are necessary for
interpreting the experiment.

The present study is a methods and calibration paper. All evaluated
controllers are short, inspectable programs with deliberately restricted
resources. Their failures are not surprising discoveries about general
reasoning. Instead, they are positive and negative controls: an instrument
that cannot distinguish a known retention deficit from a known scheduling
deficit should not be used to diagnose less interpretable systems. We now
also report exploratory evidence from one small local model, a small
Copilot-routed pilot, and a two-seed Copilot fractional factorial. None
supports general capability claims.

Our contributions are an executable fixed-work factorial task family;
an exact audit showing which structural metrics remain fixed and which
necessarily move together; and a reference-controller evaluation with
paired uncertainty estimates and analytic success predictions, followed
by controlled ablations of the assumptions underlying those profiles. We
also report one local-model factorial, paired pilot traces, and a small
fractional factorial from one Copilot-routed GPT model.
We make no claim to introduce controlled complexity evaluation itself,
to establish a universal failure threshold, or to estimate cross-model
effects from the pilot.

The distinction between task structure and controller representation is
especially important. Increasing the number of candidate slots makes
serial examination expensive, but not an indexed lookup. Increasing the
number of registers creates more values to retain, but a known deterministic
workflow can be compiled into a list of future answers. Conversely, newly
sampled unobserved values cannot be recovered by a better retention policy.
A useful structural profile must say which of these cases its controls
measure.

\section{Related Work}
\paragraph{Controlled complexity, memory demand, and tool trajectories.}
DyVal uses directed acyclic graphs to generate reasoning tasks at adjustable
complexities \cite{dyval}. Shojaee et al.\ use controlled puzzle environments
to study complexity-dependent performance and computation, including reasoning
effort under fixed inference budgets \cite{illusion}. MorphoBench adapts
question difficulty to model capability and evaluates GPT and Gemini models
\cite{morphobench}. These works establish that controllable difficulty and model-based scaling
studies are not new contributions by themselves. TRAJECT-Bench varies
parallel breadth and sequential depth in executable, production-style tool
trajectories and analyzes tool selection, argument correctness, and ordering
\cite{traject}. RHB evaluates chained tool workflows across 13 models and
tests harder variants and environmental interventions \cite{rhb}. SynthAPI
uses executable stateful environments and independently controls tool, task,
and database diversity to study training-data composition across hundreds
of thousands of tasks \cite{synthapi}. Their objectives differ from ours,
but their scope makes broad claims about the first controlled or structural
tool-use benchmark untenable.
Wang et al.\ provide an especially relevant precedent for controllable
memory difficulty: their synthetic POMDPs define memory-demand structures
and construct tunable environments while preserving the belief MDP
\cite{mds}. Our snapshot/history contrast is not a partial-observability
intervention, because successful action/update history is sufficient to
reconstruct the current registers. It measures the burden of maintaining
state internally rather than a loss of information. The hidden-resampling
extension introduces missing information, but is currently tested only with
programmed policies.
Tang et al.\ analyze depth and width in multi-agent systems \cite{tang};
their width denotes capability diversity, whereas ours counts mutable
registers. Our narrower candidate contribution is an audited, stateful
execution design that separates retention, candidate selection, and feedback
access, while measuring how controller representation changes the resulting
profile. The current experiments validate that design with programmed
controllers; they do not establish the profile of a language model.

\paragraph{Diagnostic environments.}
The Behaviour Suite explicitly evaluates core capabilities through carefully
designed tasks \cite{bsuite}. BabyAI provides synthetic instruction-following
levels of increasing difficulty and a heuristic expert \cite{babyai}.
Our reference-policy design follows this diagnostic tradition. The
workflow setting substitutes dependency-constrained operations for movement
and makes the required work and failure rules directly inspectable.
We have not experimentally compared our instrument with either platform.

\paragraph{Interactive and long-horizon evaluation.}
$\tau$-bench evaluates tool-agent-user interaction using final database states
\cite{tau}. Kwa et al.\ measure the human-duration horizons of software tasks
\cite{metr}. Those evaluation targets differ from a fixed-work intervention
on precedence. ARIES studies policies that transform interactive thought
graphs \cite{aries}; it does not provide evidence for independence of the
controls used here. The partial-observation literature distinguishes
current observations from history-conditioned information \cite{pomdp}.
That distinction is central to our observation intervention: omitted
snapshots do not destroy information already recoverable from history.

\paragraph{Tool-use benchmarks and nearest overlap.}
API-Bank, ToolBench, and AppWorld evaluate tool-augmented models on runnable
or interactive tasks, while AgentBench spans multiple interactive
environments \cite{apibank,toolbench,appworld,agentbench}. MCPAgentBench is a
particularly close recent comparison: it evaluates multi-step tool invocation
and presents candidate tool lists with distractors \cite{mcpagentbench}.
TRAJECT-Bench is a closer overlap for controlled trajectory structure and
intermediate failure analysis; RHB is a closer overlap for multi-step
tool-use difficulty and model-based experiments \cite{traject,rhb}.
MemCalib directly benchmarks model use of retrieved memory, although it
measures the influence of natural-language propositions on responses rather
than retention of mutable workflow state \cite{memcalib}. Accordingly,
neither multi-step tool use, trajectory depth, candidate selection, nor
memory evaluation is claimed as new here. Our candidate distinction is the
fixed-work state-mutating intervention and its explicit separation of
visible snapshots from history-reconstructible state. We have not
reimplemented or directly compared against these benchmark suites, and the
programmed-controller and single-model results cannot establish that this
distinction improves evaluation of language models.

\paragraph{Reasoning-budget evaluation.}
Reasoning effort and inference compute have already been studied as
experimental variables, including budget-matched comparisons of reasoning
strategies \cite{tokenbudget}. The Illusion of Thinking also studies
complexity-dependent reasoning effort \cite{illusion}. The open empirical
question for this project is not whether budget matters in general, but
whether budget changes model performance differently across audited
state-retention, candidate-selection, and feedback interventions. That
interaction has not been measured in this manuscript.

\paragraph{Workshop precedents.}
We inspected four published ALA 2025 papers as methodological precedents,
not as a systematic review or an empirical comparison. Lipkis and Agogino
study failure-probability estimation and the difficulty of sampling rare
outcomes \cite{lipkis}. Our exact small-task probabilities serve a different
purpose: they expose when an empirical zero is a sampling limitation.
Dynamic Thinker optimizes decision-time search with compute-dependent
objectives \cite{dynamic}. Our comparison counts explicit local checks and
preparation work; it does not reproduce that learned metareasoning
experiment. Nourzad et al.\ compare a task-graph scheduling architecture
with heuristic and random baselines using makespan and scheduling time
\cite{actortwin}. Our scheduler checks ordering validity, not assignment to
heterogeneous processors or makespan optimization. Leal et al.\ vary
deliberation cost in option-based learning \cite{deliberation};
our retry allowance changes execution opportunities, not option termination
or a training objective.

These precedents motivate reporting the concrete intervention, the
unrestricted reference, the resource charged, and the uncertainty of the
outcome. They do not establish that our task family is novel. The closest
methodological overlap remains controlled graph generation and diagnostic
environments; the claimed contribution is the combined audit of stateful
execution, representation-sensitive profiles, and recoverability.

\section{Executable Task Design}
\subsection{State, actions, and feedback}
A task consists of a prerequisite graph $G=(V,E)$ with $|V|=N=8$ operations,
$w\in\{1,2,4,8\}$ active registers, and a candidate count
$b\in\{2,4,8\}$. Registers take values in $\mathbb Z_8$. The initial eight-slot
array is sampled uniformly and shown once. Slots outside the active width
are unused padding. Each operation $v$ specifies a query register $q_v$,
an update register $u_v$, a nonzero increment $\delta_v\in\{1,\ldots,7\}$,
and a random permutation $\sigma_v$ of the codes $0,\ldots,b-1$.

The sole execution tool accepts an operation identifier and candidate slot:
\[
 \texttt{execute}(v,a).
\]
The action succeeds exactly when all predecessors are complete and
\[
 \sigma_v(a)=x_{q_v}\bmod b.
\]
It then marks $v$ complete and applies
$x_{u_v}\leftarrow(x_{u_v}+\delta_v)\bmod8$. Each active register is queried
exactly $N/w$ times; query positions are shuffled. Updates are sampled
independently and need not cover every register. Node labels, initial
values, update draws, and candidate-order draws are seeded.

An invalid dependency or candidate terminates the episode immediately with
an explicit failure category. Invalid tool arguments raise an error.
The goal is to complete all eight operations. Thus every successful
episode uses exactly eight execution calls. There is no reset, inspection
tool, retry, reward shaping, or recovery allowance.

Both feedback conditions reveal completed operations and the applied
register update. The \emph{snapshot} condition also repeats all current
register values. The \emph{history} condition replaces the eight values
with null entries after the initial observation. Since the initial state
and all successful updates are available, an exact history-tracking
controller can reconstruct every value in either condition. The
intervention removes convenient refreshes, not recoverable information.
It therefore cannot establish robustness to unobserved stochastic changes.

\subsection{Two fixed-work graph families}
Depth $d\in\{1,2,4,8\}$ is the longest prerequisite chain, measured in nodes.
Let $k=N/d$. The \emph{chains} family has $k$ disjoint chains of length $d$.
The \emph{barriers} family has $d$ layers of $k$ operations; every operation
in a layer depends on every operation in the preceding layer.
The two families coincide at $d=1$ and $d=N$ but differ at intermediate
depths. They are a graph-topology robustness check, not two independent
domain generators: the state and candidate mechanics are shared.

Increasing depth changes precedence while preserving $N$, the state
alphabet, candidate count, active width, and optimal execution-call count.
Holding the seed fixed pairs node labels, register draws, and candidate
orders across the snapshot intervention and, at fixed depth, across
graph families. Across width and candidate levels, the same random draws
are reused before level-specific projections. This reduces avoidable
surface variation without asserting identical semantics across levels.

\begin{table}[t]
\centering
\small
\resizebox{\columnwidth}{!}{\input{generated/audit_table}}
\caption{Exact structural audit at $N=8$. ``Masks'' counts reachable sets
of completed operations; ``orders'' counts successful operation
permutations. Ready nodes are averaged uniformly over nonterminal reachable
completion masks, not over a controller's trajectories.}
\label{tab:audit}
\end{table}

\subsection{What is and is not controlled}
We enumerate reachable completion masks and count valid schedules by
dynamic programming. For chains, the mask count is $(d+1)^k$; for barriers,
it is $d(2^k-1)+1$. The numbers of valid schedules are respectively
\[
 L_{\rm chains}=\frac{N!}{(d!)^k},
 \qquad L_{\rm barriers}=(k!)^d.
\]
Table~\ref{tab:audit} shows that depth necessarily changes reachable
completion sets and scheduling freedom. Increasing local candidate count
also multiplies the number of exposed slots across ready operations.
These are not four orthogonal graph metrics.

For a fixed initial state, commuting additive updates make the register
state a deterministic function of the completion mask. The mask audit is
therefore exact for successful executions of an instance, but excludes
failure sinks and does not claim to count a Cartesian product of all
register assignments. Across the task distribution, active initial-state
entropy is $3w$ bits; the entropy of the code tested by a candidate is
$\log_2 b$ bits. Width and candidate count consequently interact in
decision-relevant information even though the register alphabet is fixed.

The construction preserves array-slot count and the single tool interface.
It does not preserve serialized prompt length, token count, edge count,
global action branching, or total state-space size. No prompt tokens are
used by the local controllers. A causal statement in this study concerns
an operational intervention inside this generator, including its documented
couplings; it cannot be interpreted as an isolated effect of an abstract
complexity measure.

\section{Reference Controllers and Calibration}
All controllers consume public task rules and returned observations.
None can read simulator state. Except for the random scheduler, each chooses
the smallest-labelled ready operation.

The \emph{Exact} controller retains all active values and examines candidates
until it finds the matching code. \emph{Cache-2} and \emph{Cache-4} retain only
the first two or four active register values and apply public updates to
that fixed cache. A visible snapshot may answer the current query even
outside the cache. A missing value causes a uniform candidate guess.
These controllers intentionally do not infer forgotten values from their
successful guesses or precompute future answers.

\emph{Scan-1} and \emph{Scan-4} retain all registers but examine at most one
or four candidate slots. If none matches, they choose uniformly among
unexamined slots. \emph{Cache-2 + Scan-4} combines the two restrictions.
The \emph{Random schedule} controller retains all values and selects uniformly
among incomplete operations without checking dependencies. The
\emph{Uniform candidate} controller schedules correctly but guesses every
candidate. These are resource-restricted algorithms, not trained agents
or models of human cognition.

\paragraph{Analytic predictions.}
For a known queried value and a uniformly permuted candidate list, a
controller checking $s$ candidates has average per-action success
\[
 p_s=\min\{1,(s+1)/b\}.
\]
The extra one comes from the final guess. More precisely, conditional on
an instance's actual state and candidate order, success is one when the
correct slot is among those examined and $1/(b-s)$ otherwise. A query
outside the cache without a snapshot has probability $1/b$. We multiply
these instance-conditional probabilities along the deterministic ready
schedule. The initial observation is visible in both conditions, so the
first action is never incorrectly counted as a forgotten-state query.
State along a successful path is deterministic, and new uniform guesses
are independent, making this calculation exact for the specified policy.

Uniform candidate success is $b^{-N}$. The random scheduler samples a
uniform operation permutation, yielding
\[
 P_{\rm chains}=(d!)^{-k},\qquad
 P_{\rm barriers}=\frac{(k!)^d}{N!}.
\]
These formulas are calibration targets rather than fitted explanations.
In particular, critical-path depth alone cannot determine success even for
this simple scheduler: at $N=8,d=2$, its probability is $1/16$ for chains
and $1/70$ for barriers.

\section{Experimental Protocol}
The full design has $2\times4\times3\times4\times2=192$ cells.
Each uses seeds $0,\ldots,255$ and all eight controllers, for
\TaskCount{} tasks and \EpisodeCount{} episodes. The same seed is a paired
block across levels and policies. Repeated conditions are not treated as
independent trials. The configuration and analysis rules were written
before the primary run in a local protocol file; this is not an external
preregistration.

The primary outcome is complete task success. We also record completed
operations, execution calls, candidate evaluations, and first failure type.
There are no exclusions. An Exact-controller failure aborts the run.
Tests verify the full grid over additional seeds, match graph schedule
counts against exhaustive small-graph enumeration, check observation
masking and public updates, and exercise explicit invalid-action behavior.

Within each family, a main contrast is the high-minus-low endpoint
success-rate difference, averaged equally over every other level.
We report all four contrasts and a specified width-by-observation
difference-in-differences:
\begin{align*}
 I_{w,o}={}&[S(8,\text{history})-S(1,\text{history})]\\
          &-[S(8,\text{snapshot})-S(1,\text{snapshot})].
\end{align*}
The remaining controls are averaged equally in each $S$.
Percentile bootstrap bands use 2,000 resamples of the 256 paired seed blocks,
with a fixed analysis seed. They are descriptive 95\% bands, not
simultaneous significance statements. All contrast intervals and secondary
outcomes are included in the supplementary artifact.

\subsection{Separately frozen follow-on design}
After inspecting the primary results, we specified seven ablation grids
containing \ExtendedCaseCount{} task--controller configurations
(Table~\ref{tab:ablationdesign}). Each uses seeds $10{,}000,\ldots,10{,}255$,
disjoint from the primary seeds, giving \ExtendedEpisodeCount{} additional
episodes and \TotalEpisodeCount{} episodes overall. A configuration includes
its controller; the count is not a count of distinct task environments.
Some anchors deliberately recur across grids, and their repetitions do
not create independent evidence.

\begin{table*}[t]
\centering\small
\begin{tabular}{lp{.58\textwidth}r}
\toprule
Grid & Intervention and anchors & Configurations\\
\midrule
Work & $N=4,8,16$; $d=2,4$; $w=4$; initial-only or every-step snapshots & 192\\
Memory & $N=16,d=4$; fixed, next-use, residue, plan, and stale retention;
initial-only or every-fourth snapshots & 96\\
Observation & $N=8,d=4$; periods $0,1,2,4$ crossed with public updates
or hidden independent resampling & 256\\
Recovery & $N=8,d=4$; error allowances $0,2,8,56$; trials with replacement
or elimination; both snapshot endpoints & 256\\
Locality & $N=16,d=4$; balanced, clustered, or concentrated queries
crossed with additive or static state & 192\\
Selection & $N=8,d=4,w=4$; $b=2,4,8$; serial, fixed, redundant,
and indexed selection; every-step snapshots & 36\\
Labels & $N=8,d=4$; original random identifiers versus reversed identifiers;
both snapshot endpoints & 128\\
\bottomrule
\end{tabular}
\caption{Follow-on design. Both graph families appear in every grid.
Unless stated otherwise, $b=4,8$ and $w=4,8$. Grid-specific controller
sets and the complete configuration manifest are supplied with the artifact.}
\label{tab:ablationdesign}
\end{table*}

The same seed is paired across each explicitly stated comparison.
Changing $N$ necessarily changes the number of generated operations and
random draws; this is common-seed pairing, not identical task content.
Follow-on contrasts average only their stated anchors, not the complete
primary grid. All 77 selected contrasts are stored, including negative
and zero differences. The follow-on protocol was frozen locally before
these runs, but after observing the first study; neither protocol is an
external preregistration.

We additionally compute every endpoint interaction of orders two through
four in the original factorial data: six pairwise, four three-way, and
one four-way interaction for each controller and family, or 176 estimates.
These are exploratory follow-on analyses. They use signed differences
of endpoint cell means, averaging all unaffected controls equally.
Bands again resample paired seed blocks 2,000 times, with no
multiplicity-adjusted or causal-mechanism significance claims.

\section{Results}
\subsection{Distinct structural profiles}
Exact succeeds on every generated task. Cache-2 averages 75.57\% success
on chains and 75.63\% on barriers; Cache-4 averages 87.87\% and 87.89\%.
The observed Scan-1 success rates are 33.46\% and 33.52\%, whereas Scan-4
reaches 67.43\% and 67.38\% (Table~\ref{tab:summary}).
These averages reflect the chosen equal-weight grid, not a natural task
distribution.

\begin{table*}[t]
\centering
\small
\input{generated/summary_table}
\caption{Measured and instance-conditional predicted success. Intervals are
paired seed-bootstrap bands. Degenerate bootstrap bands do not bound
unseen rare events; Table~\ref{tab:scheduler} provides nondegenerate
binomial intervals for the sparse scheduling condition.}
\label{tab:summary}
\end{table*}

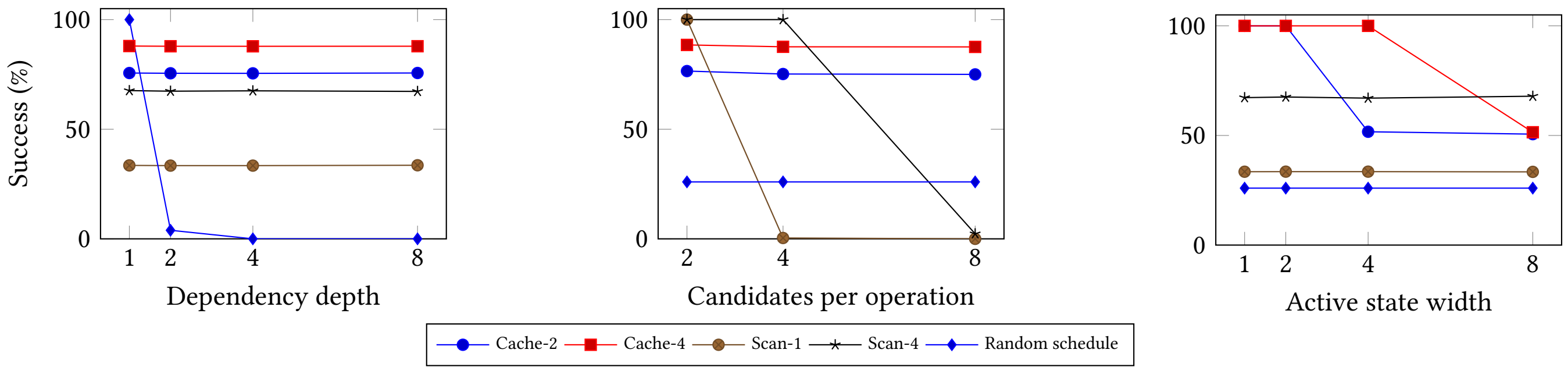
\begin{figure*}[t]
\centering
\begin{tikzpicture}
\begin{axis}[width=.31\textwidth,height=4.2cm,ymin=0,ymax=105,
  xlabel={Dependency depth},ylabel={Success (\%)},xtick={1,2,4,8},
  legend to name=profilelegend,legend columns=5,
  legend style={font=\scriptsize}]
\curve{chainsmemory2}{depth}\addlegendentry{Cache-2}
\curve{chainsmemory4}{depth}\addlegendentry{Cache-4}
\curve{chainsscan1}{depth}\addlegendentry{Scan-1}
\curve{chainsscan4}{depth}\addlegendentry{Scan-4}
\curve{chainsrandom_order}{depth}\addlegendentry{Random schedule}
\end{axis}
\end{tikzpicture}
\hfill
\begin{tikzpicture}
\begin{axis}[width=.31\textwidth,height=4.2cm,ymin=0,ymax=105,
  xlabel={Candidates per operation},xtick={2,4,8}]
\curve{chainsmemory2}{branching}
\curve{chainsmemory4}{branching}
\curve{chainsscan1}{branching}
\curve{chainsscan4}{branching}
\curve{chainsrandom_order}{branching}
\end{axis}
\end{tikzpicture}
\hfill
\begin{tikzpicture}
\begin{axis}[width=.31\textwidth,height=4.2cm,ymin=0,ymax=105,
  xlabel={Active state width},xtick={1,2,4,8}]
\curve{chainsmemory2}{width}
\curve{chainsmemory4}{width}
\curve{chainsscan1}{width}
\curve{chainsscan4}{width}
\curve{chainsrandom_order}{width}
\end{axis}
\end{tikzpicture}

\ref{profilelegend}
\caption{Marginal profiles in the chains family, averaging all other
controls equally. Points are simulation results, not fitted curves.
The width panel averages snapshot and history conditions, so its roughly
50\% cache floor hides the near-complete failure of the history-only
condition at width eight. Full curves for both families are supplied.}
\label{fig:profiles}
\Description{Three line plots show reference-controller success against
dependency depth, candidate count, and active state width. The random
scheduler is sensitive to depth, scan-limited controllers to candidate
count, and cache-limited controllers to width.}
\end{figure*}

The largest depth contrast belongs to the random scheduler: $-100$
percentage points from depth one to eight in each family.
The largest candidate contrast belongs to Scan-1, also $-100$ points.
Cache-2 instead loses approximately 49 points as width rises from one to
eight. Depth has negligible effects on the correctly scheduled cache and
scan controls compared with their resource-specific interventions.
Small residual differences reflect instance permutations and finite
simulation rather than a general equivalence claim
(Table~\ref{tab:contrasts}).

\begin{table*}[t]
\centering
\small
\input{generated/contrast_table}
\caption{High-minus-low main contrasts in percentage points. Depth: 8 minus
1; candidates: 8 minus 2; width: 8 minus 1; history: history minus snapshot.
Each averages all remaining levels equally within its family. The
supplement reports the corresponding paired intervals.}
\label{tab:contrasts}
\end{table*}

\subsection{Observation refreshes and candidate checks differ}
Both cache controllers succeed on every snapshot task, regardless of
width. Without refreshes, Cache-2 succeeds on 51.15\% of chains tasks
averaged across widths, but only about 1.2\% at width eight averaged over
depth and candidate count. Its width-by-observation interaction is
\CacheInteraction{} percentage points in chains, versus $-98.86$ in barriers.
The large interaction is expected: width creates no retention deficit when
the current value is directly visible. Averaging observation conditions
conceals this failure mode.

Increasing the scan limit changes the candidate profile but does not
recover a forgotten value. With four candidates, Scan-4 always succeeds,
while Scan-1 succeeds on fewer than 1\% of tasks. With eight candidates,
Scan-4 succeeds on 2.29\% of chains tasks and 2.15\% of barriers tasks,
consistent with $(5/8)^8\approx2.33\%$ averaged over candidate-order draws.
The combined Cache-2 + Scan-4 controller remains sensitive to both width
and candidate count. Additional local checks and increased retained state
are therefore noninterchangeable resources in this instrument.

This is not an experiment on reasoning-token budgets. Scan limits count
predicate evaluations, not tokens or elapsed computation; early failures
also shorten episodes. The follow-on redundant-check control matches
checks per decision, not total episode computation, and supports no
inference-budget conclusion.

\subsection{Calibration and rare events}
The maximum absolute discrepancy between an overall observed rate and its
analytic prediction is 0.63 percentage points, for the chains random
scheduler. Agreement is a useful implementation check, not an independent
validation of external relevance.

\begin{table}[t]
\centering
\small
\resizebox{\columnwidth}{!}{\input{generated/scheduler_table}}
\caption{Random-scheduler calibration using one representative cell per
depth and family, with 256 distinct seeds. Other candidate, width, and
observation conditions repeat the same scheduling draws and do not
increase this effective trial count. Wilson 95\% binomial intervals
retain uncertainty at zero observed successes.}
\label{tab:scheduler}
\end{table}

At barrier depth two, zero successes were observed although the exact
probability is $1/70$. With 256 independent seeded instances and scheduling
draws, this outcome has probability $(69/70)^{256}\approx2.5\%$.
The aggregate seed bootstrap then reports a degenerate 25\% band because
depth-one tasks always succeed and other sampled depths never do.
It cannot resolve an unseen rare event. The Wilson interval for the
depth-two cell extends to approximately 1.48\%, including the exact
1.43\% probability (Table~\ref{tab:scheduler}).
We therefore do not interpret a zero empirical rate or a zero-width
bootstrap band as proof of impossibility.
The Wilson intervals are a supplementary diagnostic added after observing
the sparse scheduling outcomes, not a prespecified primary analysis.

\subsection{Interactions beyond width and snapshots}
The endpoint profiles are not additive. For the combined Cache-2 + Scan-4
controller, the candidate-by-width interaction is $47.90$ percentage
points in chains and $48.14$ in barriers; the candidate-by-width-by-observation
interaction reaches $95.02$ and $95.90$ points. These large signed
differences arise partly from ceiling and floor effects: restoring a
resource can matter little while another resource already prevents
completion. They are not evidence for a universal three-factor law.
The full pairwise and higher-order tables are in the appendix, and every
interval is in the accompanying data.

\section{Ablations of the Diagnostic Assumptions}
\subsection{Work is not fixed-work depth}
The work grid increases $N$ from four to sixteen while retaining $d=2,4$,
$w=4$, and $b=4,8$. Exact succeeds throughout. Without repeated snapshots,
Cache-2 falls from 13.43\% to zero observed success: a paired difference
of $-13.43$ points with a descriptive band $[-16.02,-10.94]$.
With snapshots every step, the same cache succeeds throughout.
The scan-limited controller instead falls from 56.59\% to 50.05\%
under either snapshot endpoint. The aggregate floor near 50\% is an
artifact of averaging four-candidate tasks, which Scan-4 always solves,
with eight-candidate tasks, whose errors compound.

This is a work intervention, not a claim that dependency depth and episode
length are equivalent. More operations also create more updates and repeated
queries. The graph changes because its layer width is $N/d$; any
interpretation of the work contrast includes those effects.
The original fixed-work depth experiment and this follow-on experiment
answer different questions.

\subsection{Retention strategy and preparation}
We compare fixed caches with a \emph{next-use} cache that retains at most
two registers, prioritizing their known upcoming queries. At each snapshot,
it selects future-use registers rather than the lowest indices; a successful
guess also confirms the queried value modulo $b$. Public updates adjust the
retained residues. Its scheduling information is public, but this is a
stronger retention strategy than the deliberately non-adaptive fixed cache.

At $N=16$ with no intermediate snapshots, both limited-cache strategies
remain near the floor: Cache-2 has zero observed successes, and Next-use-2
has 0.05\%. With a snapshot every fourth successful operation,
Next-use-2 reaches \AdaptiveRefreshSuccess{}\%, compared with zero observed
success for Cache-2. The paired difference is 9.52 points
$[7.86,11.18]$. Thus a register-count profile depends on what the
controller retains and how refreshes are used; the result does not make
the two-register cache generally adequate.

Two unrestricted representation controls solve every deterministic memory
task. \emph{Residue-8} tracks each active value modulo $b$, using
$w\log_2 b$ value bits rather than $3w$. This follows because $b$ divides
eight: updates modulo eight preserve the residue needed by the candidate
predicate. \emph{Precomputed plan} uses the visible initial state and
public deterministic updates to compute all $N$ future candidate slots
before execution. It then performs $N$ plan lookups without an execution-time
register cache. The log separately records $N$ plan slots and $Nb$
candidate-table preparation units. This is neither zero memory nor free
computation: initial planning uses state workspace and the plan itself
stores information.

\subsection{Snapshot cadence versus missing information}
Snapshot periods count successful operations, not tool attempts.
Period zero means the initial observation only; period one repeats state
after every success. With public deterministic updates, Exact remains at
100\% for periods zero, one, two, and four. Cache-2 reaches 0.29\%,
100\%, 4.64\%, and 0.44\%, respectively, on the $N=8$ anchors.
The difference is therefore a retention and refresh effect, not a lack
of history-conditioned information.

We then independently resample all active registers uniformly after each
successful operation. Feedback announces that resampling occurred but
does not reveal new values unless a snapshot is due. Controllers
invalidate stale cached values. Under initial-only feedback, Exact now
has zero observed successes, despite its perfect tracking in the public-update
condition. With every-step snapshots it returns to 100\%; with period two
it achieves 0.29\%, and period four produces zero observed successes.

This case has an information bound, not merely an algorithm-specific
failure rate. Without retries, let $v=\lceil N/h\rceil$ be the number of
visible decisions at period $h>0$, and let $v=1$ for initial-only feedback.
Any history-based controller has
\[
 P(\text{complete})\le b^{-(N-v)}.
\]
At each masked decision the newly resampled target code is uniform,
independent of the entire visible history. An otherwise exact controller
attains the bound by selecting correctly at visible steps and guessing at
masked steps. For $N=8$, initial-only feedback gives $4^{-7}$ or $8^{-7}$,
not literal impossibility. The aggregate analytic probability across the
two candidate counts is approximately 0.0031\%, explaining why a modest
seed budget is unlikely to observe success.

\subsection{Recovery changes the experiment}
In the recovery grid, an invalid candidate returns its failure category
without changing state or marking an operation complete. The episode ends
only when its global error allowance $R$ is exceeded. Successful operations
still apply updates. The maximum execution horizon is $N+R$ calls.
We compare independent uniform guesses with candidate elimination, which
remembers failed codes for the current operation. This extra memory and
failure feedback must not be mistaken for additional internal reasoning.

With eight allowed errors and initial-only snapshots, uniform elimination
achieves 6.54\% success versus 1.27\% for replacement: a paired gain of
5.27 points $[4.15,6.49]$. Cache-2 with elimination reaches
\CacheRecoverySuccess{}\%. At $R=56$, elimination solves every task,
while replacement reaches 78.56\%. The average elimination cost is
\EliminateCalls{} calls, compared with eight for an unrestricted exact
controller.

The success of elimination at $R=N(b-1)$ is guaranteed, not an empirical
discovery: no operation can require more than $b-1$ incorrect trials.
For uniform replacement, the number of failed guesses before $N$ successes
is negative binomial. For elimination, failures per operation are uniform
on $\{0,\ldots,b-1\}$; convolving those counts gives the exact finite-budget
probability. The appendix states both formulas. Retry budget, error feedback,
and elimination memory together define a different task protocol from
immediate termination.

\subsection{Locality and removal of state updates}
Balanced and clustered query schedules have the same active width and
query counts. Clustering makes consecutive queries use the same register
along the deterministic ready schedule. At $N=16$ with public updates,
Next-use-2 improves from 0.05\% under balanced queries to
\ClusterSuccess{}\% under clustered queries: a paired difference of
1.81 points $[1.22,2.44]$. Reusing a confirmed residue can therefore help,
but the controller still fails most tasks. Fixed Cache-2 has zero observed
complete successes in both conditions; its progress and call counts nevertheless
differ. A completion-only score misses that difference.

Concentrating every query on register zero yields 100\% for both limited
cache strategies. This is a deliberate \emph{invalid width control}:
the declared width remains four or eight, but the effective queried width
is one. We report that audit failure instead of presenting concentration
as a fair improvement at fixed memory demand.

The \emph{Stale-8} controller retains initial values but ignores updates.
Removing updates restores it from zero observed success on balanced
additive tasks to 100\% on balanced static tasks; unrestricted tracking
remains at 100\% in both. This negative control confirms that the update
mechanism matters. Static tasks are easier for this stale strategy, not
evidence that tracking mutable state is an intrinsic universal difficulty.

\subsection{Candidate indexing and redundant checks}
\begin{table*}[t]
\centering\small
\input{generated/selection_resources}
\caption{Selection ablation with full snapshots. Checks per call are the
aggregate comparison count divided by execution calls. Build entries
charge candidate-table preparation separately; lookup calls are separate
operations, not equivalent CPU-time units. Early termination changes
total comparisons, so only the fixed and redundant controls match
comparison counts per decision.}
\label{tab:selectionresources}
\end{table*}

The \emph{Fixed-check-4} controller evaluates four candidates per decision
and uses every result. The \emph{Sham-check-4} controller also evaluates
four, but only the first result is allowed to affect selection; the other
three are redundant. With four candidates, Fixed-check-4 succeeds at
100\% and Sham-check-4 at 0.59\%. With eight candidates, they reach
1.17\% and zero observed success. Sham-check-4 produces exactly the same
action traces as Scan-1 under paired seeds, despite its higher comparison
cost. Useful computation, rather than charged computation alone, explains
the difference.

\emph{Indexed} constructs code-to-slot maps before execution and retains
all required state. It succeeds at 100\% for all tested candidate counts,
using eight execution lookups and $Nb$ preparation entries
(Table~\ref{tab:selectionresources}). Branching is therefore costly for the
serial-scan controller but not an inherent obstacle to this representation.
Index preparation grows with $b$, and we do not compare total wall time,
energy, or token costs. The experiment is a counterexample to
representation-free interpretation of a branching profile, not proof that
indexing is costless.

\subsection{Identifiers, floors, and aggregation}
Reversing the randomly assigned operation identifiers preserves the graph
audit and Exact's 100\% success. It changes the deterministic ready order
and hence some queried states. Scan-4's paired difference is $-0.24$ points
$[-0.59,0.15]$ with initial-only feedback, and Cache-2's is $-0.15$
$[-0.49,0.15]$. This limited relabeling check does not establish invariance
to arbitrary wording, prompts, or tool names. Random scheduling remains
too sparse at the selected depth to resolve subtle differences.

Several ablations are dominated by ceiling and floor effects. We retain
zero-result conditions, exact probabilities where derived, and secondary
progress and resource outcomes rather than regenerate tasks after observing
success. The follow-on grid is consequently useful for checking assumptions
but not calibrated for ranking unfamiliar agents. Aggregate percentages
must always be accompanied by the conditions averaged into them.

\section{Interpretation and Limitations}
The experiment shows that the instrument responds selectively to
limitations introduced by construction. It also supplies a concrete
counterexample to using depth alone as a universal difficulty measure:
equal-depth, equal-work graphs can have different scheduling probabilities.
For an unfamiliar controller, these profiles would be diagnostic evidence,
not a unique identification of its internal mechanism.

Several limitations constrain the claims. The tasks are small and
mechanical; the single execution tool omits natural-language instructions,
tool discovery, and uncertain user goals. Immediate termination in the
primary protocol
magnifies compounding error and is a design choice, not a typical feature
of software workflows. Even the follow-on maximum of sixteen operations
does not represent long real-world workflows.

Active register count is not a lower bound on a controller's required
memory. A controller could retain values modulo $b$, compress state, or
precompute future answers, as the new controls explicitly demonstrate.
The restricted-cache results measure chosen strategies, not minimum
memory requirements. The hidden-resampling extension is one extreme
missing-information mechanism, not a general model of partial observation,
ambiguous feedback, or hidden user intent.

The graph families share the same state mechanics and are identical at the
endpoint depths. The comparison is limited topology replication, not
independent-domain generalization. Main contrasts include documented
couplings, and the equal-weight grid is arbitrary. The large rollout count
must not be confused with an equally large number of independent samples.
All endpoint interactions are reported, but their bootstrap bands are
descriptive and can degenerate at rare-event floors. No fitted universal
difficulty model or out-of-domain predictive validation is supplied.

The next-use controller receives the same public graph as other controllers,
but exploits future query knowledge and learns residues from successful
trials. Its cache entry count does not include graph metadata, readiness
bookkeeping, or controller code. Likewise, the residue-bit counts cover
stored values, not total implementation memory. Preparation entries,
lookups, and predicate checks are separate operation categories;
none is a hardware-neutral measure of compute.

Language-model evidence is exploratory and limited to one Copilot-routed
model pilot, a 16-episode half-fraction over two task seeds, and one small
local-model factorial with a success floor. The Copilot pilot uses three
paired task seeds per intervention cell, with its third seed added after
inspection of the first two. The fractional factorial's last episode was
completed after the first 15 outcomes had been inspected, though its
prespecified condition was unchanged. These studies use one routed model
and one quantized 0.5B model; none includes a reasoning-budget comparison,
cross-provider comparison, or transfer to real tasks. They cannot establish
which current model breaks first, whether more reasoning helps, or how the
profiles predict real-task performance.

\section{Exploratory Language-Model Pilot}
At each decision, the model received the complete public task description,
current observation, and interaction history, then returned one JSON action.
The environment validated and executed the action; an invalid candidate or
dependency choice ended the task. This is a serialized text-action
interface, not native provider function calling. Copilot Auto selected the
model, recorded as GPT-6 Luna throughout. The account did not expose a
selectable Gemini or Qwen route or a reasoning-effort condition. This pilot
therefore tests one model route, not providers or inference-budget scaling.

The pilot covers four paired low/high interventions in the chain family,
with eight operations per task and three task seeds per cell. The first two
seeds (16 episodes) were run before a third paired seed (eight episodes)
was added; the latter was chosen after inspecting the initial outcomes. We
therefore report only descriptive counts, without significance tests or
generalization claims. Table \ref{tab:modelpilot} reports the first
completed run for each unique factor--level--seed cell. The raw log contains
29 records, including five repeated/recovery records excluded from the
24-episode summaries. Success was 2/3 versus
3/3 for depth 1 versus 8, 2/3 versus 3/3 for branching 2 versus 8, 2/3
versus 2/3 for register width 1 versus 8, and 0/3 versus 2/3 for history
versus snapshot feedback. These tiny counts are compatible with substantial
sampling uncertainty; they do not establish that depth helps or that any
factor has a stable model-level effect.

\begin{table}[t]
\centering\small
\input{results_model_pilot/pilot_summary}
\caption{Exploratory Copilot-routed GPT-6 Luna pilot. Mean completed
operations are out of eight. Each cell has three task seeds; no inferential
claims are warranted.}
\label{tab:modelpilot}
\end{table}

\subsection{Two-seed fractional-factorial follow-up}
We completed a second, separately frozen Copilot Auto study using the same
text-action interface and GPT-6 Luna route. It covers eight conditions in a
$2^{4-1}$ resolution-IV half fraction of depth (2 or 8), branching (2 or
8), register width (1 or 8), and history or snapshot feedback; every task
requires eight operations. Two task seeds yield 16 episodes. The design is
balanced and orthogonal for its main-effect contrasts, but main effects
remain aliased with three-factor interactions, and two-factor interactions
are mutually aliased. The two-seed sample is too small for reliable
uncertainty estimates.

The model completed 8/16 workflows and 88/128 required operations (5.5 per
task on average). Seven episodes ended in candidate-selection errors and
one in a dependency error. The descriptive high-minus-low success
differences were $-25$, $+50$, $+25$, and $+50$ percentage points for
depth, branching, width, and snapshot feedback, respectively; mean
completed-operation differences were $-1.50$, $+1.50$, $+2.50$, and
$+2.00$. Table \ref{tab:model-factorial} summarizes these raw contrasts.
The signs, including the apparent benefit of branching, are not a basis for
mechanistic interpretation: there are only two seeds, possible
three-factor aliasing, and a single model route. The last of 16 episodes
was run after the first 15 outcomes had been inspected following an
interruption; its condition and seed had already been specified, and the
design was not changed. This exploratory result does not rescue the
floor-level local-model factorial or establish a reproducible agent
capability profile.

\begin{table}[t]
\centering\small
\input{generated/model_factorial_summary}
\caption{Descriptive high-minus-low contrasts in the 16-episode
Copilot-routed fractional factorial. Eight episodes contribute to each
factor level; each of only two task seeds spans four conditions per level.
Main effects may be aliased with three-factor interactions.}
\label{tab:model-factorial}
\end{table}

\section{Exploratory Local-Model Factorial}
We ran a complete $2^5$ factorial of 256 tasks using the local,
quantized Qwen2.5-0.5B-Instruct model \cite{qwen}. The five controls were
graph family (chains or barriers), dependency depth (2 or 8), candidate
count (2 or 8), active-register width (1 or 8), and feedback format (history
or current-state snapshot). Every task required eight operations. Eight
task seeds were paired across all 32 cells. The protocol, factor grid,
seeds, and analysis were fixed before this planned run; the protocol is
local rather than externally preregistered.

The model received the complete public task specification, current
observation, and interaction history, and returned one JSON action per
decision. Invalid JSON or an invalid action ended the episode without a
retry. Inference used llama.cpp server b12030 on CPU, Q4\_K\_M weights,
temperature zero, and a 64-token output limit \cite{llamacpp}. No hosted
provider or paid inference was used. This is a single, small-model
evaluation, not a comparison of providers or reasoning budgets. The
separate runtime smoke test was excluded from the factorial and is retained
in the artifact.

Across the planned tasks, the model completed
\LocalSuccessCount/\LocalEpisodeCount episodes
(\LocalOverallSuccessRate{} success rate) and completed only
\LocalCompletedOperationCount{} of \LocalRequiredOperationCount{} required
operations (\LocalMeanCompleted{} of eight per task on average). Of
\LocalDecisionCount{} decisions,
\LocalSchemaValidCount{} (\LocalSchemaValidRate{}) were strictly
schema-valid JSON. The first-failure counts were 162 dependency errors, 54
candidate-selection errors, 18 invalid actions, 16 invalid action schemas,
and 6 invalid JSON outputs. Median decision latency was
\LocalMedianLatency{} seconds on the local CPU.

Table \ref{tab:localmodel} reports marginal high-minus-low success
differences and seed-cluster bootstrap intervals. All are zero because
every condition had zero completed workflows. These degenerate intervals
are a floor effect, not evidence that the structural factors have no
effect; the model was insufficiently capable on this interface to resolve
their effects on the success endpoint. The run therefore documents a
failed calibration of this model-task pairing, not a useful structural
capability profile. It cannot establish generalization to larger or hosted
models, and it does not test whether additional reasoning budget changes
the profile.

\begin{table*}[t]
\centering\small
\input{generated/local_model_summary}
\caption{Exploratory local-model factorial. Each level contains 128 tasks.
Differences are paired by task seed and averaged over the other factors;
intervals resample the eight seed clusters.}
\label{tab:localmodel}
\end{table*}

\section{Next Empirical Test}
The next study should freeze the interaction harness and task protocol before
inference; use larger paired samples; and compare multiple models and
providers. It should record model version, every prompt and action, valid
action rate, completed operations, token usage, provider-reported cost, and
latency. A reasoning-budget intervention must use a supported, explicit
budget control and a total-usage-matched comparison. Retries and extra tool
calls must remain separate interventions. An independent state-transition
family and realistic transfer tasks are necessary before claiming
generality. The present account has a 24-episode pilot and a 16-episode,
two-seed fractional factorial using one Auto-routed model, plus a local
small-model run at the success floor. Provider-level, cross-model, and
reasoning-budget comparisons remain uncompleted.

\section{Conclusion}
Structural task controls are useful only when their invariants and
couplings are explicit. Our fixed-work workflow instrument distinguishes
scheduling, candidate-search, and retention restrictions with executable
controls and exact calibration targets. The reference experiments show
why a structural profile is more interpretable than its aggregate score,
while also showing why depth, state width, and observation access cannot
be treated as universally independent difficulty variables.

The local-model factorial is a failed calibration of the model-task
interface, not evidence that the instrument yields a generalizable profile:
one quantized 0.5B model completed none of 256 workflows, creating a floor
on the success endpoint. The completed-operation and action-error records
are useful diagnostics, but the zero marginal success contrasts cannot be
interpreted as structural invariance. A second, 16-episode Copilot study
shows nonzero contrasts, but two task seeds and fractional-factorial
aliasing make them unstable descriptive observations, not evidence of
dimension-specific effects. Claims about larger models, provider
differences, reasoning-budget effects, or transfer to real workflows remain
open.

\paragraph{Reproducibility and research integrity.}
The accompanying artifact contains both reference-task generators, the
model-pilot, fractional-factorial, and local-model harnesses, 35 tests,
protocols, episode traces,
provider-usage records, structural audits, analysis scripts, generated
tables, and source hashes. The 24 unique Copilot-routed model episodes use
three paired seeds per intervention cell; the third seed was added after
inspection of the first two, and five repeated/recovery records are excluded
from their summaries. The Copilot fractional factorial has 16 episodes,
two task seeds across eight planned half-fraction conditions, and 96 logged
model calls; the final episode was completed after inspecting
the first 15 outcomes, without changing the prespecified design. The
separate local factorial uses one quantized
Qwen2.5-0.5B model, eight paired seeds, and 256 planned episodes; it uses no
paid inference. Copilot usage logs report 264 premium-request units and
25.653 billion nano-AIU across the recorded smoke and pilot runs; these are
usage units, not dollars, and the account-wide quota is unavailable. There
are no human-subject data or external-system actions. The code and experiment records are available at
\url{https://github.com/Shereen20112003/what-structurally-breaks-an-agent}.
Model weights and third-party binaries are not included in that repository.

\ifdefined\shortversion
\else
\clearpage
\onecolumn
\appendix
\input{appendix_body}
\fi
\end{document}

%% file: generated/numbers.tex
\newcommand{\TaskCount}{49,152}
\newcommand{\EpisodeCount}{393,216}

\newcommand{\CacheInteraction}{-98.80}

%% file: generated/curve_data.tex
\expandafter\def\csname chainsmemory2depth\endcsname{(1,75.6348) (2,75.5208) (4,75.4883) (8,75.6510)}
\expandafter\def\csname chainsmemory2branching\endcsname{(2,76.5259) (4,75.1831) (8,75.0122)}
\expandafter\def\csname chainsmemory2width\endcsname{(1,100.0000) (2,100.0000) (4,51.6927) (8,50.6022)}
\expandafter\def\csname chainsmemory4depth\endcsname{(1,87.9395) (2,87.8418) (4,87.8255) (8,87.8581)}
\expandafter\def\csname chainsmemory4branching\endcsname{(2,88.4521) (4,87.5977) (8,87.5488)}
\expandafter\def\csname chainsmemory4width\endcsname{(1,100.0000) (2,100.0000) (4,100.0000) (8,51.4648)}
\expandafter\def\csname chainsscan1depth\endcsname{(1,33.4961) (2,33.3984) (4,33.3984) (8,33.5612)}
\expandafter\def\csname chainsscan1branching\endcsname{(2,100.0000) (4,0.3906) (8,0.0000)}
\expandafter\def\csname chainsscan1width\endcsname{(1,33.4635) (2,33.4961) (4,33.4961) (8,33.3984)}
\expandafter\def\csname chainsscan4depth\endcsname{(1,67.6107) (2,67.3503) (4,67.5130) (8,67.2526)}
\expandafter\def\csname chainsscan4branching\endcsname{(2,100.0000) (4,100.0000) (8,2.2949)}
\expandafter\def\csname chainsscan4width\endcsname{(1,67.2526) (2,67.5456) (4,67.0247) (8,67.9036)}
\expandafter\def\csname chainsrandom_orderdepth\endcsname{(1,100.0000) (2,3.9062) (4,0.0000) (8,0.0000)}
\expandafter\def\csname chainsrandom_orderbranching\endcsname{(2,25.9766) (4,25.9766) (8,25.9766)}
\expandafter\def\csname chainsrandom_orderwidth\endcsname{(1,25.9766) (2,25.9766) (4,25.9766) (8,25.9766)}
\expandafter\def\csname barriersmemory2depth\endcsname{(1,75.6348) (2,75.5208) (4,75.6999) (8,75.6510)}
\expandafter\def\csname barriersmemory2branching\endcsname{(2,76.6968) (4,75.1709) (8,75.0122)}
\expandafter\def\csname barriersmemory2width\endcsname{(1,100.0000) (2,100.0000) (4,51.9368) (8,50.5697)}
\expandafter\def\csname barriersmemory4depth\endcsname{(1,87.9395) (2,87.8743) (4,87.9069) (8,87.8581)}
\expandafter\def\csname barriersmemory4branching\endcsname{(2,88.4521) (4,87.6709) (8,87.5610)}
\expandafter\def\csname barriersmemory4width\endcsname{(1,100.0000) (2,100.0000) (4,100.0000) (8,51.5788)}
\expandafter\def\csname barriersscan1depth\endcsname{(1,33.4961) (2,33.4961) (4,33.5286) (8,33.5612)}
\expandafter\def\csname barriersscan1branching\endcsname{(2,100.0000) (4,0.5615) (8,0.0000)}
\expandafter\def\csname barriersscan1width\endcsname{(1,33.6263) (2,33.5286) (4,33.5286) (8,33.3984)}
\expandafter\def\csname barriersscan4depth\endcsname{(1,67.6107) (2,67.3503) (4,67.3177) (8,67.2526)}
\expandafter\def\csname barriersscan4branching\endcsname{(2,100.0000) (4,100.0000) (8,2.1484)}
\expandafter\def\csname barriersscan4width\endcsname{(1,67.3503) (2,67.4805) (4,66.9922) (8,67.7083)}
\expandafter\def\csname barriersrandom_orderdepth\endcsname{(1,100.0000) (2,0.0000) (4,0.0000) (8,0.0000)}
\expandafter\def\csname barriersrandom_orderbranching\endcsname{(2,25.0000) (4,25.0000) (8,25.0000)}
\expandafter\def\csname barriersrandom_orderwidth\endcsname{(1,25.0000) (2,25.0000) (4,25.0000) (8,25.0000)}

%% file: generated/extended_numbers.tex
\newcommand{\ExtendedEpisodeCount}{295,936}
\newcommand{\ExtendedCaseCount}{1,156}
\newcommand{\TotalEpisodeCount}{689,152}
\newcommand{\AdaptiveRefreshSuccess}{9.52}
\newcommand{\CacheRecoverySuccess}{44.63}
\newcommand{\EliminateCalls}{27.87}
\newcommand{\ClusterSuccess}{1.86}

%% file: generated/local_model_numbers.tex
\newcommand{\LocalEpisodeCount}{256}
\newcommand{\LocalSuccessCount}{0}
\newcommand{\LocalOverallSuccessRate}{0.000}
\newcommand{\LocalCompletedOperationCount}{21}
\newcommand{\LocalRequiredOperationCount}{2048}
\newcommand{\LocalMeanCompleted}{0.08}
\newcommand{\LocalDecisionCount}{277}
\newcommand{\LocalSchemaValidCount}{255}
\newcommand{\LocalSchemaValidRate}{0.921}
\newcommand{\LocalMedianLatency}{9.00}

%% file: generated/audit_table.tex
\begin{tabular}{lrrrrr}
\toprule
Family & $d$ & Edges & Masks & Orders & Ready nodes\\
\midrule
Chains & 1 & 0 & 256 & 40320 & 4.02\\
Chains & 2 & 4 & 81 & 2520 & 2.70\\
Chains & 4 & 6 & 25 & 70 & 1.67\\
Chains & 8 & 7 & 9 & 1 & 1.00\\
Barriers & 1 & 0 & 256 & 40320 & 4.02\\
Barriers & 2 & 16 & 31 & 576 & 2.13\\
Barriers & 4 & 12 & 13 & 16 & 1.33\\
Barriers & 8 & 7 & 9 & 1 & 1.00\\
\bottomrule
\end{tabular}

%% file: generated/summary_table.tex
\begin{tabular}{llrrr}
\toprule
Family & Controller & Success (\%) & 95\% interval & Predicted (\%)\\
\midrule
Barriers & Exact & 100.00 & [100.00, 100.00] & 100.00\\
Barriers & Cache-2 & 75.63 & [75.51, 75.75] & 75.56\\
Barriers & Cache-2 + Scan-4 & 51.06 & [50.87, 51.27] & 51.13\\
Barriers & Cache-4 & 87.89 & [87.80, 88.00] & 87.93\\
Barriers & Random schedule & 25.00 & [25.00, 25.00] & 25.37\\
Barriers & Scan-1 & 33.52 & [33.45, 33.61] & 33.46\\
Barriers & Scan-4 & 67.38 & [67.20, 67.56] & 67.42\\
Barriers & Uniform candidate & 0.11 & [0.06, 0.18] & 0.13\\
Chains & Exact & 100.00 & [100.00, 100.00] & 100.00\\
Chains & Cache-2 & 75.57 & [75.46, 75.68] & 75.56\\
Chains & Cache-2 + Scan-4 & 51.18 & [51.00, 51.38] & 51.18\\
Chains & Cache-4 & 87.87 & [87.78, 87.96] & 87.93\\
Chains & Random schedule & 25.98 & [25.39, 26.66] & 26.61\\
Chains & Scan-1 & 33.46 & [33.40, 33.54] & 33.45\\
Chains & Scan-4 & 67.43 & [67.26, 67.64] & 67.49\\
Chains & Uniform candidate & 0.14 & [0.07, 0.21] & 0.13\\
\bottomrule
\end{tabular}

%% file: generated/contrast_table.tex
\begin{tabular}{llrrrr}
\toprule
Family & Controller & Depth & Candidates & Width & History\\
\midrule
Barriers & Exact & 0.00 & 0.00 & 0.00 & 0.00\\
Barriers & Cache-2 & 0.02 & -1.68 & -49.43 & -48.75\\
Barriers & Cache-2 + Scan-4 & -0.10 & -74.79 & -33.32 & -32.63\\
Barriers & Cache-4 & -0.08 & -0.89 & -48.42 & -24.21\\
Barriers & Random schedule & -100.00 & 0.00 & 0.00 & 0.00\\
Barriers & Scan-1 & 0.07 & -100.00 & -0.23 & 0.00\\
Barriers & Scan-4 & -0.36 & -97.85 & 0.36 & 0.00\\
Barriers & Uniform candidate & 0.13 & -0.34 & 0.13 & 0.00\\
Chains & Exact & 0.00 & 0.00 & 0.00 & 0.00\\
Chains & Cache-2 & 0.02 & -1.51 & -49.40 & -48.85\\
Chains & Cache-2 + Scan-4 & -0.10 & -74.49 & -33.32 & -32.69\\
Chains & Cache-4 & -0.08 & -0.90 & -48.54 & -24.27\\
Chains & Random schedule & -100.00 & 0.00 & 0.00 & 0.00\\
Chains & Scan-1 & 0.07 & -100.00 & -0.07 & 0.00\\
Chains & Scan-4 & -0.36 & -97.71 & 0.65 & 0.00\\
Chains & Uniform candidate & 0.13 & -0.42 & 0.00 & 0.00\\
\bottomrule
\end{tabular}

%% file: generated/scheduler_table.tex
\begin{tabular}{lrrrr}
\toprule
Family & $d$ & Observed (\%) & Wilson interval & Exact (\%)\\
\midrule
Chains & 1 & 100.00 & [98.52, 100.00] & 100.0000\\
Chains & 2 & 3.91 & [2.14, 7.04] & 6.2500\\
Chains & 4 & 0.00 & [0.00, 1.48] & 0.1736\\
Chains & 8 & 0.00 & [0.00, 1.48] & 0.0025\\
Barriers & 1 & 100.00 & [98.52, 100.00] & 100.0000\\
Barriers & 2 & 0.00 & [0.00, 1.48] & 1.4286\\
Barriers & 4 & 0.00 & [0.00, 1.48] & 0.0397\\
Barriers & 8 & 0.00 & [0.00, 1.48] & 0.0025\\
\bottomrule
\end{tabular}

%% file: generated/selection_resources.tex
\begin{tabular}{rlrrrr}
\toprule
$b$ & Controller & Success (\%) & Checks/call & Build entries & Lookups\\
\midrule
4 & Exact & 100.00 & 2.49 & 0 & 0.00\\
4 & Indexed & 100.00 & 0.00 & 32 & 8.00\\
4 & Scan-1 & 0.59 & 1.00 & 0 & 0.00\\
4 & Sham-check-4 & 0.59 & 4.00 & 0 & 0.00\\
4 & Fixed-check-4 & 100.00 & 4.00 & 0 & 0.00\\
8 & Exact & 100.00 & 4.52 & 0 & 0.00\\
8 & Indexed & 100.00 & 0.00 & 64 & 8.00\\
8 & Scan-1 & 0.00 & 1.00 & 0 & 0.00\\
8 & Sham-check-4 & 0.00 & 4.00 & 0 & 0.00\\
8 & Fixed-check-4 & 1.17 & 4.00 & 0 & 0.00\\
\bottomrule
\end{tabular}

%% file: results_model_pilot/pilot_summary.tex
\begin{tabular}{llrrr}
\toprule
Intervention & Level & Success & Mean completed & $n$ \\
\midrule
Depth & 1 & 2/3 & 5.67 & 3 \\
Depth & 8 & 3/3 & 8.00 & 3 \\
Branching & 2 & 2/3 & 7.33 & 3 \\
Branching & 8 & 3/3 & 8.00 & 3 \\
Register width & 1 & 2/3 & 7.33 & 3 \\
Register width & 8 & 2/3 & 6.33 & 3 \\
Feedback & History & 0/3 & 5.00 & 3 \\
Feedback & Snapshot & 2/3 & 5.33 & 3 \\
\bottomrule
\end{tabular}

%% file: generated/model_factorial_summary.tex
\begin{tabular}{lrrr}
\toprule
Factor & Success difference & Operation difference & $n$ per level \\
\midrule
Depth (8 vs. 2) & -25.0 pp & -1.50 & 8 \\
Branching (8 vs. 2) & +50.0 pp & +1.50 & 8 \\
Register width (8 vs. 1) & +25.0 pp & +2.50 & 8 \\
Snapshot vs. history & +50.0 pp & +2.00 & 8 \\
\bottomrule
\end{tabular}

%% file: generated/local_model_summary.tex
\begin{tabular}{lrrr}
\toprule
Contrast & Low success & High success & Difference (95\% interval) \\
\midrule
family (barriers $-$ chains) & 0/128 (0.000) & 0/128 (0.000) & +0.0 pp (+0.0, +0.0) \\
depth (8 $-$ 2) & 0/128 (0.000) & 0/128 (0.000) & +0.0 pp (+0.0, +0.0) \\
branching (8 $-$ 2) & 0/128 (0.000) & 0/128 (0.000) & +0.0 pp (+0.0, +0.0) \\
width (8 $-$ 1) & 0/128 (0.000) & 0/128 (0.000) & +0.0 pp (+0.0, +0.0) \\
observation (snapshot $-$ history) & 0/128 (0.000) & 0/128 (0.000) & +0.0 pp (+0.0, +0.0) \\
\bottomrule
\end{tabular}

%% file: appendix_body.tex
\section{Complete Experimental Specification}
\label{app:specification}
The primary experiment and follow-on experiment are separately recorded.
The primary grid uses $N=8$, depths $\{1,2,4,8\}$, candidate counts
$\{2,4,8\}$, active widths $\{1,2,4,8\}$, both snapshot endpoints, and
both graph families. Its seeds are 0--255 and each task is evaluated by
eight programmed controllers. The follow-on grids use the anchors
in the main-text ablation design; their full per-controller configurations
are in \path{results_extended/case_design.json}.

\paragraph{Pairing and randomness.}
The base generator draws a fixed eight-slot initial array, a permutation
of operation identifiers, a shuffled balanced query assignment, update
registers, nonzero increments, and candidate permutations. Candidate-count
levels filter the same eight-code permutation; active-width levels project
query and update indices into the active width. Across work levels, different
numbers of operations cause different subsequent random draws, so seed
pairing is not exact surface matching. Hidden resampling uses a separate
seeded random stream, draws eight fresh values after each success, and
replaces only active registers. Failed trials never advance that stream.
Controller sampling uses its own stream. Common controller seeds pair
variants, but differing controller logic can consume different numbers
of draws. These choices are intended for reproducibility, not to certify
statistical independence of every repeated condition.

\paragraph{Admission and positive controls.}
There is no outcome-based rejection or regeneration. All balanced tasks
must query every active register; clustered assignments preserve that
coverage and its multiplicities. Concentrated-query variants explicitly
have effective width one and are tagged as a degenerate manipulation.
Positive controls include unrestricted deterministic tracking, residue
tracking, deterministic preplanning, unrestricted tracking with a snapshot
after every hidden resampling, and candidate elimination with the worst-case
error allowance. The runner aborts if a designated positive control fails,
if the call horizon is exceeded, or if the adaptive cache exceeds two
entries. Analytic probabilities are derived only for supported cases;
unsupported predictions are labelled \texttt{not\_derived}, not replaced
with a default or fitted estimate.

\paragraph{Adaptive-cache details.}
Next-use-2 uses the public deterministic ready schedule to prioritize
registers queried in the future. On a visible snapshot, the current query
can be answered directly, so future-use selection skips that operation.
During history-only feedback, a successful action confirms the queried
residue from the chosen candidate code. The controller inserts that
residue, applies the public increment if relevant, and evicts values
whose next use is furthest away. This is not a generic learned memory
strategy; future-query metadata, successful-action feedback, and public
updates are deliberate advantages of the control.

\paragraph{Preparation and storage accounting.}
Residue caches store values modulo $b$; raw-value caches store values
modulo eight. Reported value bits exclude dictionary keys, graph metadata,
controller code, temporary operands, and runtime overhead. Preplanning
records $N$ candidate slots, using the initial state as preparation
workspace; indexed selection records $N$ maps of $b$ code-to-slot entries.
The preparation counter charges $Nb$ table entries for each of these
representations. It is a workload proxy, not a measurement of exact executed
CPU instructions or comparison count in Python's \texttt{index} function.
Lookup counts are kept separate from predicate comparisons. The experiment
does not aggregate these categories into an invented universal compute unit.

\section{Calibration Derivations}
\label{app:proofs}
\subsection{Completion states and schedule probabilities}
For $k=N/d$ disjoint chains, a completion set is determined by the completed
prefix length $0,\ldots,d$ of each chain, giving $(d+1)^k$ reachable masks.
Each chain imposes one internal order on its $d$ operations; interleaving
the $k$ orders gives $N!/(d!)^k$ valid permutations. A uniformly sampled
operation permutation therefore succeeds with probability $(d!)^{-k}$.

For barriers, all previous layers must be complete before any operation
in the next layer succeeds. Each of the $d$ layers contributes $2^k-1$
nonempty partial-completion choices, with one initial empty mask, giving
$d(2^k-1)+1$ distinct masks. Each layer's operations can be internally
permuted in $k!$ ways, so the valid-order count is $(k!)^d$ and random-order
success is $(k!)^d/N!$. Failed-action sinks are not included in these
completion-mask counts.

\subsection{Deterministic history and residue sufficiency}
Let $C_t$ be the completed-operation set after $t$ successes. Because each
update is additive and public,
\[
 x_r(t)=x_r(0)+
 \sum_{v\in C_t:u_v=r}\delta_v \pmod8.
\]
Thus the initial observation and successful-action history determine every
register exactly. Repeating snapshots cannot improve optimal information
availability in this deterministic protocol, although it can help a
restricted retention strategy. Since each tested $b$ divides eight, the
same equation modulo $b$ determines every candidate predicate. Tracking
$x_r\bmod b$ is therefore sufficient; storing all three raw value bits
is not necessary for $b=2$ or $b=4$.

A known ready schedule can also be simulated once using the initial
observation. Recording its future candidate slots is an exact plan in the
deterministic protocol. That construction establishes an alternative
representation, not a lower bound on memory and not a claim that planning
uses less total storage or preparation.

\subsection{Missing-information ceiling}
In the hidden-resampling condition, all active values are freshly uniform
after each success. Before a masked decision, conditioning on any visible
history still leaves the queried code uniform on $0,\ldots,b-1$.
Choosing a different ready operation cannot change this fact, because every
active register has the same independent fresh distribution. Consequently,
the conditional probability of each hidden decision succeeding is at most
$1/b$, regardless of additional computation.

With no retries, the number of visible decisions on a successful path is
$v=\lceil N/h\rceil$ for snapshot period $h>0$, or $v=1$ for initial-only
feedback. Applying the chain rule gives $P(\text{complete})\le b^{-(N-v)}$.
An unrestricted exact controller attains this ceiling. The bound does not
apply unchanged to recovery protocols: failed trials preserve state and
provide information that can support elimination. For scan-limited
controllers, formulas that average visible candidate-order draws are
ensemble predictions and are labelled separately from exact
instance-conditional predictions.

\subsection{Finite retry budgets}
Uniform replacement guesses have success probability $p=1/b$ per attempt.
If $F$ is the number of errors before $N$ successes, then
\[
 \Pr(F\le R)=\sum_{r=0}^{R}
   {N+r-1\choose r}p^N(1-p)^r.
\]
Under uniform elimination, the correct code has a uniformly random rank
in a permutation of the $b$ candidates. Errors per operation are therefore
independent uniform counts on $\{0,\ldots,b-1\}$. The finite-budget
probability is
\[
 \Pr(F\le R)=\frac{1}{b^N}
 \sum_{r=0}^{R}[z^r](1+z+\cdots+z^{b-1})^N.
\]
The implementation evaluates this coefficient sum by a truncated
convolution. If $R\ge N(b-1)$ the probability is one. Mean calls without
budget exhaustion are $Nb$ for replacement and $N(b+1)/2$ for elimination.
These are exact properties of the selected guessing policies, not
general guarantees for arbitrary tool-using controllers.

\section{All Follow-On Outcome Summaries}
\label{app:ablationtables}
Each row below averages equally over the remaining anchors of its grid.
Low and high are descriptive 95\% percentile bands from 2,000 paired
seed-block bootstrap resamples. They are not simultaneous confidence
statements. Constants produce degenerate bands; a sampled zero is not
a proof of zero probability. Calls include failed attempts and stop at
termination. For exact analytic predictions, progress, error counts,
comparison counts, preparation workload, lookups, and cache measurements,
see \path{results_extended/summary.csv}. All 77 paired contrast estimates
and their bands are in \path{results_extended/paired_contrasts.csv}.

Condition notation uses \texttt{n} for operation count, \texttt{p} for
snapshot period, \texttt{r} for global error allowance, and
\texttt{shock1} for independent hidden resampling. Period zero means
initial-only state feedback. The \texttt{reverse} flag indicates operation
identifier reversal. All grids use both graph families.

\small
\subsection{Operation Count}
\input{generated/ablation_work}
\subsection{Retention Strategy}
\input{generated/ablation_memory}
\subsection{Snapshot Cadence and Hidden Resampling}
\input{generated/ablation_observation}
\subsection{Recovery and Candidate Elimination}
\input{generated/ablation_recovery}
\subsection{Query Locality and State Updates}
\input{generated/ablation_locality}
\subsection{Candidate Selection and Resource Controls}
\input{generated/ablation_selection}
\subsection{Operation Identifiers}
\input{generated/ablation_labels}
\normalsize

\section{All Endpoint Interaction Estimates}
\label{app:interactions}
The following estimates are exploratory follow-on analyses of the original
192-cell factorial experiment. $D,B,W,O$ denote depth, candidate count,
active width, and history-only versus snapshot feedback. Endpoint
contrasts use high minus low for each participating dimension and average
all nonparticipating controls over their original levels. A pairwise
interaction is a difference-in-differences; the corresponding signed
construction generalizes to orders three and four. Values are percentage
points, so an interaction is not itself a success probability. Per-estimate
bootstrap bands are retained in
\path{results_extended/all_endpoint_interactions.csv}.

\begin{center}
\small
\resizebox{\textwidth}{!}{\input{generated/pairwise_interactions}}

\medskip
\resizebox{\textwidth}{!}{\input{generated/higher_interactions}}
\end{center}

\section{Workshop Comparison and Submission Scope}
We inspected four ALA 2025 papers as a purposive sample, not all workshop
papers and not a systematic novelty search. Their total PDF lengths are
nine pages (failure sampling), ten (Dynamic Thinker), ten (task-graph
scheduling), and eight (deliberation cost). References begin on pages
nine, eight, seven, and eight, respectively. Total PDF length therefore
does not equal the allowed main-text length.

The historical ALA 2025 call allows original work up to eight pages
excluding references, and preliminary or outlook papers up to six pages.
It specifies double-blind review and deanonymized camera-ready copies.
Our expanded authored version is a preprint-style manuscript with an
appendix; the separate \path{review.tex} source removes the author block
and appendix, but must still be checked against the current workshop's
track and page rules. A longer document is not evidence of acceptance
or of a stronger contribution. No current-edition acceptance, conference
metadata, or submission date is asserted.

\section{Reproduction and Evidence Boundaries}
The benchmark, controllers, tests, and analysis use the Python standard
library. Document inspection additionally uses the pinned PDF packages.
From the project directory in a Windows PowerShell terminal:
\begin{verbatim}
.\.venv\Scripts\python.exe -m unittest -v test_benchmark test_extended
.\.venv\Scripts\python.exe run_experiments.py --seeds 256
.\.venv\Scripts\python.exe analyze.py
.\.venv\Scripts\python.exe run_extended.py
.\.venv\Scripts\python.exe analyze_extended.py
.\build.ps1
.\.venv\Scripts\python.exe verify_artifact.py
.\.venv\Scripts\python.exe package_artifact.py
\end{verbatim}
Re-running the primary experiment replaces its generated records;
the source and protocol hashes must match the supplied metadata when
reproducing that study. The original and follow-on episode files have
separate hashes and execution metadata. The two locally frozen protocols
and the time ordering of exploratory analyses are retained explicitly.
Complete primary contrasts, rare-scheduler Wilson intervals, and example
traces remain available in the original result directory.

This artifact contains only programmed-controller evidence. No frontier
model, learned policy, training run, provider reasoning-budget intervention,
human participant, or realistic transfer task has been evaluated.
Source compilation and local reproducibility do not establish arXiv
moderation approval, workshop acceptance, or compatibility with a
particular submission processor. The author must review the manuscript
and current submission rules before uploading it.

%% file: generated/ablation_work.tex
\begin{longtable}{llrrrr}
\toprule
Condition & Controller & Success (\%) & Low & High & Calls\\
\midrule
\endhead
n16\_p0 & Cache-2 & 0.00 & 0.00 & 0.00 & 3.38\\
n16\_p0 & Exact & 100.00 & 100.00 & 100.00 & 16.00\\
n16\_p0 & Random schedule & 0.00 & 0.00 & 0.00 & 1.69\\
n16\_p0 & Scan-4 & 50.05 & 50.00 & 50.15 & 9.44\\
n16\_p1 & Cache-2 & 100.00 & 100.00 & 100.00 & 16.00\\
n16\_p1 & Exact & 100.00 & 100.00 & 100.00 & 16.00\\
n16\_p1 & Random schedule & 0.00 & 0.00 & 0.00 & 1.69\\
n16\_p1 & Scan-4 & 50.05 & 50.00 & 50.15 & 9.44\\
n4\_p0 & Cache-2 & 13.43 & 10.94 & 16.21 & 2.93\\
n4\_p0 & Exact & 100.00 & 100.00 & 100.00 & 4.00\\
n4\_p0 & Random schedule & 10.94 & 8.11 & 13.96 & 1.60\\
n4\_p0 & Scan-4 & 56.59 & 54.88 & 58.50 & 3.13\\
n4\_p1 & Cache-2 & 100.00 & 100.00 & 100.00 & 4.00\\
n4\_p1 & Exact & 100.00 & 100.00 & 100.00 & 4.00\\
n4\_p1 & Random schedule & 10.94 & 8.01 & 13.96 & 1.60\\
n4\_p1 & Scan-4 & 56.59 & 54.88 & 58.40 & 3.13\\
n8\_p0 & Cache-2 & 0.73 & 0.29 & 1.22 & 3.17\\
n8\_p0 & Exact & 100.00 & 100.00 & 100.00 & 8.00\\
n8\_p0 & Random schedule & 1.17 & 0.49 & 1.95 & 1.63\\
n8\_p0 & Scan-4 & 50.59 & 50.24 & 51.03 & 5.18\\
n8\_p1 & Cache-2 & 100.00 & 100.00 & 100.00 & 8.00\\
n8\_p1 & Exact & 100.00 & 100.00 & 100.00 & 8.00\\
n8\_p1 & Random schedule & 1.17 & 0.49 & 1.95 & 1.63\\
n8\_p1 & Scan-4 & 50.59 & 50.20 & 51.03 & 5.18\\
\bottomrule
\end{longtable}

%% file: generated/ablation_memory.tex
\begin{longtable}{llrrrr}
\toprule
Condition & Controller & Success (\%) & Low & High & Calls\\
\midrule
\endhead
p0 & Next-use-2 & 0.05 & 0.00 & 0.15 & 5.05\\
p0 & Cache-2 & 0.00 & 0.00 & 0.00 & 2.98\\
p0 & Cache-4 & 50.00 & 50.00 & 50.00 & 9.66\\
p0 & Residue-8 & 100.00 & 100.00 & 100.00 & 16.00\\
p0 & Precomputed plan & 100.00 & 100.00 & 100.00 & 16.00\\
p0 & Stale-8 & 0.00 & 0.00 & 0.00 & 4.20\\
p4 & Next-use-2 & 9.52 & 7.86 & 11.28 & 7.68\\
p4 & Cache-2 & 0.00 & 0.00 & 0.00 & 3.10\\
p4 & Cache-4 & 50.00 & 50.00 & 50.00 & 9.76\\
p4 & Residue-8 & 100.00 & 100.00 & 100.00 & 16.00\\
p4 & Precomputed plan & 100.00 & 100.00 & 100.00 & 16.00\\
p4 & Stale-8 & 2.98 & 2.05 & 4.05 & 5.33\\
\bottomrule
\end{longtable}

%% file: generated/ablation_observation.tex
\begin{longtable}{llrrrr}
\toprule
Condition & Controller & Success (\%) & Low & High & Calls\\
\midrule
\endhead
p0\_shock0 & Cache-2 & 0.29 & 0.05 & 0.59 & 2.85\\
p0\_shock0 & Exact & 100.00 & 100.00 & 100.00 & 8.00\\
p0\_shock0 & Scan-4 & 50.68 & 50.34 & 51.12 & 5.18\\
p0\_shock0 & Uniform & 0.00 & 0.00 & 0.00 & 1.25\\
p0\_shock1 & Cache-2 & 0.00 & 0.00 & 0.00 & 2.25\\
p0\_shock1 & Exact & 0.00 & 0.00 & 0.00 & 2.25\\
p0\_shock1 & Scan-4 & 0.00 & 0.00 & 0.00 & 2.01\\
p0\_shock1 & Uniform & 0.00 & 0.00 & 0.00 & 1.25\\
p1\_shock0 & Cache-2 & 100.00 & 100.00 & 100.00 & 8.00\\
p1\_shock0 & Exact & 100.00 & 100.00 & 100.00 & 8.00\\
p1\_shock0 & Scan-4 & 50.68 & 50.34 & 51.12 & 5.18\\
p1\_shock0 & Uniform & 0.00 & 0.00 & 0.00 & 1.25\\
p1\_shock1 & Cache-2 & 100.00 & 100.00 & 100.00 & 8.00\\
p1\_shock1 & Exact & 100.00 & 100.00 & 100.00 & 8.00\\
p1\_shock1 & Scan-4 & 50.98 & 50.54 & 51.46 & 5.25\\
p1\_shock1 & Uniform & 0.00 & 0.00 & 0.00 & 1.25\\
p2\_shock0 & Cache-2 & 4.64 & 3.66 & 5.71 & 3.58\\
p2\_shock0 & Exact & 100.00 & 100.00 & 100.00 & 8.00\\
p2\_shock0 & Scan-4 & 50.68 & 50.34 & 51.17 & 5.18\\
p2\_shock0 & Uniform & 0.00 & 0.00 & 0.00 & 1.25\\
p2\_shock1 & Cache-2 & 0.29 & 0.05 & 0.59 & 2.49\\
p2\_shock1 & Exact & 0.29 & 0.05 & 0.59 & 2.49\\
p2\_shock1 & Scan-4 & 0.24 & 0.05 & 0.54 & 2.21\\
p2\_shock1 & Uniform & 0.00 & 0.00 & 0.00 & 1.25\\
p4\_shock0 & Cache-2 & 0.44 & 0.20 & 0.73 & 2.93\\
p4\_shock0 & Exact & 100.00 & 100.00 & 100.00 & 8.00\\
p4\_shock0 & Scan-4 & 50.68 & 50.29 & 51.07 & 5.18\\
p4\_shock0 & Uniform & 0.00 & 0.00 & 0.00 & 1.25\\
p4\_shock1 & Cache-2 & 0.00 & 0.00 & 0.00 & 2.27\\
p4\_shock1 & Exact & 0.00 & 0.00 & 0.00 & 2.27\\
p4\_shock1 & Scan-4 & 0.00 & 0.00 & 0.00 & 2.03\\
p4\_shock1 & Uniform & 0.00 & 0.00 & 0.00 & 1.25\\
\bottomrule
\end{longtable}

%% file: generated/ablation_recovery.tex
\begin{longtable}{llrrrr}
\toprule
Condition & Controller & Success (\%) & Low & High & Calls\\
\midrule
\endhead
r0\_p0 & Cache-2 + eliminate & 0.29 & 0.05 & 0.59 & 2.85\\
r0\_p0 & Scan-1 + eliminate & 0.29 & 0.10 & 0.54 & 1.64\\
r0\_p0 & Uniform & 0.00 & 0.00 & 0.00 & 1.25\\
r0\_p0 & Eliminate & 0.00 & 0.00 & 0.00 & 1.25\\
r0\_p1 & Cache-2 + eliminate & 100.00 & 100.00 & 100.00 & 8.00\\
r0\_p1 & Scan-1 + eliminate & 0.29 & 0.10 & 0.54 & 1.64\\
r0\_p1 & Uniform & 0.00 & 0.00 & 0.00 & 1.25\\
r0\_p1 & Eliminate & 0.00 & 0.00 & 0.00 & 1.25\\
r2\_p0 & Cache-2 + eliminate & 3.52 & 2.59 & 4.49 & 5.86\\
r2\_p0 & Scan-1 + eliminate & 2.93 & 2.10 & 3.76 & 5.11\\
r2\_p0 & Uniform & 0.05 & 0.00 & 0.15 & 3.72\\
r2\_p0 & Eliminate & 0.00 & 0.00 & 0.00 & 3.92\\
r2\_p1 & Cache-2 + eliminate & 100.00 & 100.00 & 100.00 & 8.00\\
r2\_p1 & Scan-1 + eliminate & 2.93 & 2.10 & 3.81 & 5.11\\
r2\_p1 & Uniform & 0.05 & 0.00 & 0.15 & 3.72\\
r2\_p1 & Eliminate & 0.00 & 0.00 & 0.00 & 3.92\\
r56\_p0 & Cache-2 + eliminate & 100.00 & 100.00 & 100.00 & 18.94\\
r56\_p0 & Scan-1 + eliminate & 100.00 & 100.00 & 100.00 & 21.52\\
r56\_p0 & Uniform & 78.56 & 76.61 & 80.52 & 43.27\\
r56\_p0 & Eliminate & 100.00 & 100.00 & 100.00 & 27.87\\
r56\_p1 & Cache-2 + eliminate & 100.00 & 100.00 & 100.00 & 8.00\\
r56\_p1 & Scan-1 + eliminate & 100.00 & 100.00 & 100.00 & 21.52\\
r56\_p1 & Uniform & 78.56 & 76.71 & 80.47 & 43.27\\
r56\_p1 & Eliminate & 100.00 & 100.00 & 100.00 & 27.87\\
r8\_p0 & Cache-2 + eliminate & 44.63 & 42.77 & 46.39 & 13.63\\
r8\_p0 & Scan-1 + eliminate & 43.07 & 41.75 & 44.38 & 12.81\\
r8\_p0 & Uniform & 1.27 & 0.78 & 1.81 & 11.09\\
r8\_p0 & Eliminate & 6.54 & 5.42 & 7.67 & 12.64\\
r8\_p1 & Cache-2 + eliminate & 100.00 & 100.00 & 100.00 & 8.00\\
r8\_p1 & Scan-1 + eliminate & 43.07 & 41.75 & 44.43 & 12.81\\
r8\_p1 & Uniform & 1.27 & 0.78 & 1.81 & 11.09\\
r8\_p1 & Eliminate & 6.54 & 5.42 & 7.67 & 12.64\\
\bottomrule
\end{longtable}

%% file: generated/ablation_locality.tex
\begin{longtable}{llrrrr}
\toprule
Condition & Controller & Success (\%) & Low & High & Calls\\
\midrule
\endhead
balanced\_additive & Next-use-2 & 0.05 & 0.00 & 0.15 & 5.05\\
balanced\_additive & Cache-2 & 0.00 & 0.00 & 0.00 & 2.98\\
balanced\_additive & Exact & 100.00 & 100.00 & 100.00 & 16.00\\
balanced\_additive & Stale-8 & 0.00 & 0.00 & 0.00 & 4.20\\
balanced\_static & Next-use-2 & 0.05 & 0.00 & 0.15 & 5.05\\
balanced\_static & Cache-2 & 0.00 & 0.00 & 0.00 & 3.04\\
balanced\_static & Exact & 100.00 & 100.00 & 100.00 & 16.00\\
balanced\_static & Stale-8 & 100.00 & 100.00 & 100.00 & 16.00\\
clustered\_additive & Next-use-2 & 1.86 & 1.32 & 2.44 & 7.67\\
clustered\_additive & Cache-2 & 0.00 & 0.00 & 0.00 & 7.26\\
clustered\_additive & Exact & 100.00 & 100.00 & 100.00 & 16.00\\
clustered\_additive & Stale-8 & 0.15 & 0.00 & 0.39 & 4.80\\
clustered\_static & Next-use-2 & 1.76 & 1.17 & 2.34 & 7.68\\
clustered\_static & Cache-2 & 0.00 & 0.00 & 0.00 & 7.24\\
clustered\_static & Exact & 100.00 & 100.00 & 100.00 & 16.00\\
clustered\_static & Stale-8 & 100.00 & 100.00 & 100.00 & 16.00\\
concentrated\_additive & Next-use-2 & 100.00 & 100.00 & 100.00 & 16.00\\
concentrated\_additive & Cache-2 & 100.00 & 100.00 & 100.00 & 16.00\\
concentrated\_additive & Exact & 100.00 & 100.00 & 100.00 & 16.00\\
concentrated\_additive & Stale-8 & 9.42 & 6.93 & 12.11 & 7.21\\
concentrated\_static & Next-use-2 & 100.00 & 100.00 & 100.00 & 16.00\\
concentrated\_static & Cache-2 & 100.00 & 100.00 & 100.00 & 16.00\\
concentrated\_static & Exact & 100.00 & 100.00 & 100.00 & 16.00\\
concentrated\_static & Stale-8 & 100.00 & 100.00 & 100.00 & 16.00\\
\bottomrule
\end{longtable}

%% file: generated/ablation_selection.tex
\begin{longtable}{llrrrr}
\toprule
Condition & Controller & Success (\%) & Low & High & Calls\\
\midrule
\endhead
b2 & Exact & 100.00 & 100.00 & 100.00 & 8.00\\
b2 & Indexed & 100.00 & 100.00 & 100.00 & 8.00\\
b2 & Scan-1 & 100.00 & 100.00 & 100.00 & 8.00\\
b2 & Sham-check-4 & 100.00 & 100.00 & 100.00 & 8.00\\
b2 & Scan-2 & 100.00 & 100.00 & 100.00 & 8.00\\
b2 & Fixed-check-4 & 100.00 & 100.00 & 100.00 & 8.00\\
b4 & Exact & 100.00 & 100.00 & 100.00 & 8.00\\
b4 & Indexed & 100.00 & 100.00 & 100.00 & 8.00\\
b4 & Scan-1 & 0.59 & 0.00 & 1.37 & 2.00\\
b4 & Sham-check-4 & 0.59 & 0.00 & 1.37 & 2.00\\
b4 & Scan-2 & 9.18 & 6.64 & 12.11 & 3.55\\
b4 & Fixed-check-4 & 100.00 & 100.00 & 100.00 & 8.00\\
b8 & Exact & 100.00 & 100.00 & 100.00 & 8.00\\
b8 & Indexed & 100.00 & 100.00 & 100.00 & 8.00\\
b8 & Scan-1 & 0.00 & 0.00 & 0.00 & 1.25\\
b8 & Sham-check-4 & 0.00 & 0.00 & 0.00 & 1.25\\
b8 & Scan-2 & 0.00 & 0.00 & 0.00 & 1.61\\
b8 & Fixed-check-4 & 1.17 & 0.20 & 2.34 & 2.39\\
\bottomrule
\end{longtable}

%% file: generated/ablation_labels.tex
\begin{longtable}{llrrrr}
\toprule
Condition & Controller & Success (\%) & Low & High & Calls\\
\midrule
\endhead
reverse0\_p0 & Cache-2 & 0.29 & 0.05 & 0.59 & 2.85\\
reverse0\_p0 & Exact & 100.00 & 100.00 & 100.00 & 8.00\\
reverse0\_p0 & Random schedule & 0.00 & 0.00 & 0.00 & 1.29\\
reverse0\_p0 & Scan-4 & 50.68 & 50.34 & 51.12 & 5.18\\
reverse0\_p1 & Cache-2 & 100.00 & 100.00 & 100.00 & 8.00\\
reverse0\_p1 & Exact & 100.00 & 100.00 & 100.00 & 8.00\\
reverse0\_p1 & Random schedule & 0.00 & 0.00 & 0.00 & 1.29\\
reverse0\_p1 & Scan-4 & 50.68 & 50.29 & 51.12 & 5.18\\
reverse1\_p0 & Cache-2 & 0.15 & 0.00 & 0.34 & 2.88\\
reverse1\_p0 & Exact & 100.00 & 100.00 & 100.00 & 8.00\\
reverse1\_p0 & Random schedule & 0.00 & 0.00 & 0.00 & 1.35\\
reverse1\_p0 & Scan-4 & 50.44 & 50.15 & 50.83 & 5.23\\
reverse1\_p1 & Cache-2 & 100.00 & 100.00 & 100.00 & 8.00\\
reverse1\_p1 & Exact & 100.00 & 100.00 & 100.00 & 8.00\\
reverse1\_p1 & Random schedule & 0.00 & 0.00 & 0.00 & 1.35\\
reverse1\_p1 & Scan-4 & 50.44 & 50.10 & 50.83 & 5.23\\
\bottomrule
\end{longtable}

%% file: generated/pairwise_interactions.tex
\begin{tabular}{llrrrrrr}
\toprule
Family & Controller & $D\times{} B$ & $D\times{} W$ & $D\times{} O$ & $B\times{} W$ & $B\times{} O$ & $W\times{} O$\\
\midrule
Barriers & exact & 0.00 & 0.00 & 0.00 & 0.00 & 0.00 & 0.00\\
Barriers & memory2 & -0.10 & 0.10 & 0.10 & -2.05 & -2.05 & -97.95\\
Barriers & memory2\_scan4 & 0.49 & -1.27 & 0.10 & 48.93 & 47.56 & -50.68\\
Barriers & memory4 & 0.68 & -0.68 & -0.68 & -4.00 & -4.00 & -95.61\\
Barriers & random\_order & 0.00 & 0.00 & 0.00 & 0.00 & 0.00 & 0.00\\
Barriers & scan1 & 0.00 & 0.00 & 0.00 & 0.00 & 0.00 & 0.00\\
Barriers & scan4 & -0.59 & 0.59 & 0.00 & 2.93 & 0.00 & 0.00\\
Barriers & uniform & -0.59 & 0.98 & 0.00 & -0.59 & 0.00 & 0.00\\
Chains & exact & 0.00 & 0.00 & 0.00 & 0.00 & 0.00 & 0.00\\
Chains & memory2 & -0.10 & 0.10 & 0.10 & -2.05 & -2.05 & -97.95\\
Chains & memory2\_scan4 & 0.49 & -1.27 & 0.10 & 48.93 & 47.56 & -50.68\\
Chains & memory4 & 0.68 & -0.68 & -0.68 & -4.00 & -4.00 & -95.61\\
Chains & random\_order & 0.00 & 0.00 & 0.00 & 0.00 & 0.00 & 0.00\\
Chains & scan1 & 0.00 & 0.00 & 0.00 & 0.00 & 0.00 & 0.00\\
Chains & scan4 & -0.59 & 0.59 & 0.00 & 2.93 & 0.00 & 0.00\\
Chains & uniform & -0.59 & 0.98 & 0.00 & -0.59 & 0.00 & 0.00\\
\bottomrule
\end{tabular}

%% file: generated/higher_interactions.tex
\begin{tabular}{llrrrrr}
\toprule
Family & Controller & $D\times{} B\times{} W$ & $D\times{} B\times{} O$ & $D\times{} W\times{} O$ & $B\times{} W\times{} O$ & $D\times{} B\times{} W\times{} O$\\
\midrule
Barriers & exact & 0.00 & 0.00 & 0.00 & 0.00 & 0.00\\
Barriers & memory2 & -0.20 & -0.20 & 0.20 & -4.10 & -0.39\\
Barriers & memory2\_scan4 & -1.37 & 1.37 & 0.20 & 95.12 & 2.73\\
Barriers & memory4 & 1.37 & 1.37 & -1.37 & -8.01 & 2.73\\
Barriers & random\_order & 0.00 & 0.00 & 0.00 & 0.00 & 0.00\\
Barriers & scan1 & 0.00 & 0.00 & 0.00 & 0.00 & 0.00\\
Barriers & scan4 & 1.17 & 0.00 & 0.00 & 0.00 & 0.00\\
Barriers & uniform & -1.95 & 0.00 & 0.00 & 0.00 & 0.00\\
Chains & exact & 0.00 & 0.00 & 0.00 & 0.00 & 0.00\\
Chains & memory2 & -0.20 & -0.20 & 0.20 & -4.10 & -0.39\\
Chains & memory2\_scan4 & -1.37 & 1.37 & 0.20 & 95.12 & 2.73\\
Chains & memory4 & 1.37 & 1.37 & -1.37 & -8.01 & 2.73\\
Chains & random\_order & 0.00 & 0.00 & 0.00 & 0.00 & 0.00\\
Chains & scan1 & 0.00 & 0.00 & 0.00 & 0.00 & 0.00\\
Chains & scan4 & 1.17 & 0.00 & 0.00 & 0.00 & 0.00\\
Chains & uniform & -1.95 & 0.00 & 0.00 & 0.00 & 0.00\\
\bottomrule
\end{tabular}